\documentclass[sigconf]{acmart}

\usepackage{booktabs} 
\usepackage{bm}
\usepackage{multirow}
\usepackage{xspace}
\usepackage[tight,footnotesize]{subfigure}
\usepackage{algorithm}   
\usepackage{amsmath}
\usepackage{enumitem}
\usepackage{algorithmic}
\usepackage{balance}
\usepackage{cancel}
\usepackage{CJKutf8}
\usepackage[T1]{fontenc}
\usepackage[utf8]{inputenc}
\usepackage{microtype}
\usepackage{graphicx}
\usepackage{bbding}
\usepackage{float}
\usepackage{colortbl}
\usepackage{longtable}

\newcommand{\paratitle}[1]{
\noindent\textbf{#1}}
\newcommand{\ie}{\emph{i.e.,}\xspace}

\newcommand{\eg}{\emph{e.g.,}\xspace}

\newcommand{\wrt}{\emph{w.r.t.}\xspace}

\newcommand{\model}{\emph{PAIR}\xspace}
\newcommand{\kg}{\emph{MoKG}\xspace}
\newcommand{\lightmodel}{\emph{LightPAIR}\xspace}

\setcopyright{acmcopyright}
\copyrightyear{2018}
\acmYear{2018}
\acmDOI{XXXXXXX.XXXXXXX}

\acmConference[Conference acronym 'XX]{Make sure to enter the correct
  conference title from your rights confirmation email}{June 03--05,
  2018}{Woodstock, NY}
\acmPrice{15.00}
\acmISBN{978-1-4503-XXXX-X/18/06}

\begin{document}

\title[]{Making Large Language Models Better Knowledge Miners for Online Marketing with Progressive Prompting Augmentation}

\author{Chunjing Gan}
\authornotemark[1]
\affiliation{
\institution{Ant Group}
\country{}
}
\email{cuibing.gcj@antgroup.com}

\author{Dan Yang}
\authornotemark[1]
\affiliation{
\institution{Ant Group}
\country{}
}
\email{luoyin.yd@antgroup.com}
\thanks{* Equal contribution.}

\author{Binbin Hu}
\authornotemark[1]
\affiliation{
\institution{Ant Group}
\country{}
}
\email{bin.hbb@antgroup.com}

\author{Ziqi Liu}
\affiliation{
\institution{Ant Group}
\country{}
}
\email{ziqiliu@antgroup.com}

\author{Yue Shen}
\affiliation{
\institution{Ant Group}
\country{}
}
\email{zhanying@antgroup.com}

\author{Zhiqiang Zhang}
\affiliation{
\institution{Ant Group}
\country{}
}
\email{lingyao.zzq@antgroup.com}

\author{Jinjie Gu}
\affiliation{
\institution{Ant Group}
\country{}
}
\email{jinjie.gujj@antgroup.com}

\author{Jun Zhou}
\authornotemark[2]
\affiliation{
\institution{Ant Group}
\country{}
}
\email{jun.zhoujun@antgroup.com}
\thanks{$^{\dagger}$ Corresponding author.}

\author{Guannan Zhang}
\affiliation{
\institution{Ant Group}
\country{}
}
\email{zgn138592@antgroup.com}

\begin{abstract}
Nowadays, the rapid development of mobile economy has promoted the flourishing of online marketing campaigns, whose success greatly hinges on the efficient matching between user preferences and desired marketing campaigns where a well-established Marketing-oriented Knowledge Graph (dubbed as \kg) could serve as the critical ``bridge'' for preference propagation.
In this paper, we seek to carefully prompt a Large Language Model (LLM) with domain-level knowledge as a better marketing-oriented knowledge miner for marketing-oriented knowledge graph construction, which is however non-trivial, suffering from several inevitable issues in real-world marketing scenarios, \ie  uncontrollable relation generation of LLMs, insufficient prompting ability of a single prompt, the unaffordable deployment cost of LLMs. To this end, we propose  {\model},  a novel \underline{P}rogressive prompting \underline{A}ugmented m\underline{I}ning f\underline{R}amework for harvesting marketing-oriented knowledge graph with LLMs. 
In particular, we reduce the pure relation generation to an LLM based adaptive relation filtering process through the knowledge-empowered prompting technique.
Next, we steer LLMs for entity expansion with progressive prompting augmentation, followed by a reliable aggregation with comprehensive consideration of both self-consistency and semantic relatedness.
In terms of online serving, we specialize in a small and white-box {\model} (\ie {\lightmodel}), which is fine-tuned with a high-quality corpus provided by a strong teacher-LLM. 
Extensive experiments and practical applications in audience targeting verify the effectiveness of the proposed (\emph{Light}){\model}~\footnote{Source code and datasets will be released upon acceptance.}.
\end{abstract}



\maketitle

\section{Introduction}
\begin{figure}[t]
    \centering
    \includegraphics[width=1.0\columnwidth]{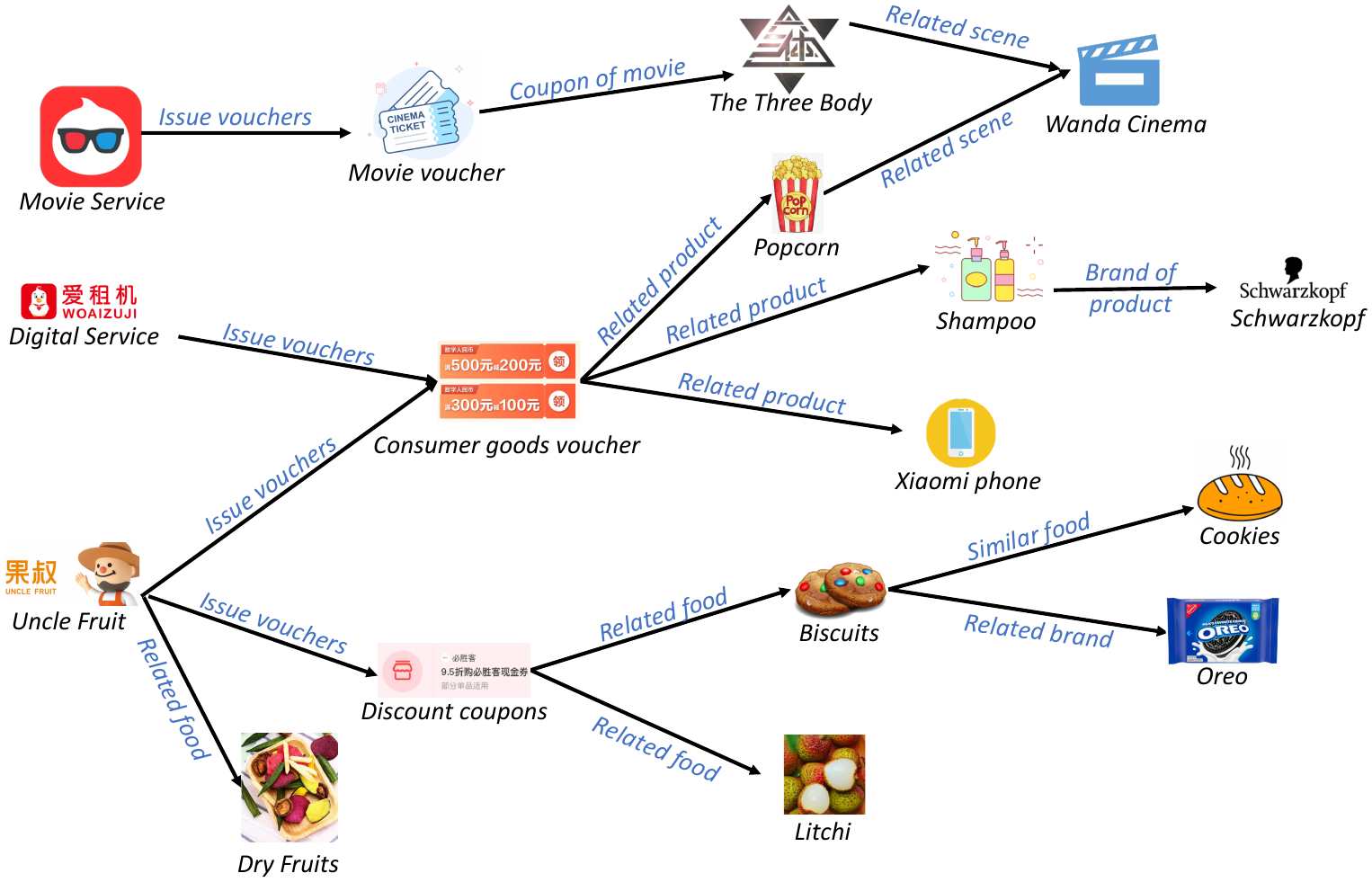}
    \caption{An example of the marketing-oriented knowledge graph.}
    \label{fig:kg}
\end{figure}

Nowadays, the rapid growth of mobile economy has established the prosperity of online marketing campaigns~\cite{marketing2021_1,marketing2021_2,marketing2023_1,marketing2023_2}, which attracts a variety of Internet companies or merchants to promote their products / services. 
Taking Alipay as an example, as an integrated service platform in China, numerous merchants conduct marketing campaigns (\eg offering coupons, vouchers, and bonuses) through so-called mini-apps each day~\cite{zhuang2020hubble,campaign2021_1}, in which efficient information delivery plays a crucially important role.  Fundamentally, the core is to perform intelligent matching between user preference and desired campaigns\cite{campaign2019_1,campaign2020,campaign2023_1,campaign2023_2}, where a well-established Marketing-oriented Knowledge Graph (dubbed as \kg) could serve as the critical ``bridge'', allowing flexible and effective preference propagation. Although SupKG~\cite{zang2023supkg}\footnote{SupKG involves more than ten millions
of entities, more than one hundred relations, and more than one hundred million of triplets.} has made considerable efforts in connecting users and various services in Alipay, product-related entities with hierarchical and  spatiotemporal relations are at its heart, while {\kg} emphasizes more on the marketing-related relations between users and merchants (A snapshot of {\kg} is illustrated in Fig.~\ref{fig:kg}),
as a kind of complementary knowledge to enrich the original SupKG graph (More details in Sec.~\ref{exp:casestudy2}).

Intuitively, we could directly graft the construction pipeline of SupKG, which is a mixture of multiple text mining techniques (\eg NER~\cite{ner2018_1,ner2020_1,ner2020_2,ner2022_1,ner2023_1} and RE~\cite{Zhu2019re,re2023_1,re2023_2,re2023_3}), to supplement marketing-related entities and relations.  
Unfortunately, its success greatly hinges on manual labeling and evaluation, which is rather expensive. 
On the other hand, Large Language Models (LLMs) (\eg  ChatGPT~\cite{chatgpt2023}, GPT4~\cite{gpt42023}, Llama~
\cite{touvron2023llama1,touvron2023llama2}, ChatGLM~\cite{du2022glm,zeng2023glmb}) have shown remarkable and even superhuman capabilities on various NLP tasks.
Learning from a vast corpus of Internet text, LLMs have encoded a huge amount of open-world knowledge~\cite{lmaskb2019,lmaskb2022,crawling2023}, which are naturally excellent helpers to harvest factual information for knowledge graphs~\cite{bertnet2023,crawling2023,zhu2023llms}. 
However, the effectiveness of LLM based mining approaches cannot be always guaranteed in real-world marketing scenarios, since pure LLMs may be unfamiliar with the marketing-oriented entities and relations\cite{tail2023,cov2023}.
Thus, \emph{can we bridge both worlds,} \ie \emph{carefully prompting LLMs with prior knowledge (e.g., SupKG or other knowledge) as a better marketing-oriented knowledge miner ?}

With this in mind, we decompose the whole mining process of {\kg} as the following three stages: 
 \emph{Prior Knowledge} $\rightarrow$ \emph{Relation Generation}  $\rightarrow$ \emph{Entity Expansion}.
 With the help of prompting engineering, domain-level prior knowledge could be easily injected into LLMs for relation generation and entity expansion, whereas the performance is still far from optimal for real-world marketing-oriented scenarios due to the following unresolved issues:
 (i) \emph{Pure relation generation is uncontrollable.} 
Despite the impressive performance of LLMs in various generative tasks, it is undeniable that the capabilities of LLMs can get out of control very quickly in relation exploration. In other words, LLMs may inevitably generate 
several
useless / fallacious~\cite{zhang2023hallucination,Zhang2023HowLMhallucinate,jiang2023active} relations for given entities (\eg Take the entity ``The Three Body'' in Fig.~\ref{fig:kg} as an example, the LLM would tell several undesired relations towards marketing, \eg ``Related scientists'' and ``Related organizations''.) 
so that the following entity expansion would be greatly confused. 
 (ii) \emph{A single prompt is insufficient.} 
Intuitively, the performance of entity expansion depends highly on the quality of the prompt used to steer the LLMs. Unfortunately, due to the variety of prior knowledge, coupled with the sensitivity of prompting under slight modifications~\cite{sensitivity2020_1,sensitivity2023_1,sensitivity2023_2}, a single painstakingly crafted prompt may not always produce the desired results.
 (iii) \emph{A white-box and smaller model is preferred.} 
Generally, real-world marketing scenarios involve millions of seed entities, which brings far beyond affordable cost to directly serve (or call open API of) a performant LLM (\eg GPT 3.5). On the other hand, the growing concerns for privacy demand to circumvent a black-box LLM in realistic applications.

To address the above problems, we propose a novel \underline{P}rogressive prompting \underline{A}ugmented m\underline{I}ning f\underline{R}amework for harvesting marketing-oriented knowledge with LLMs (dubbed as {\model}).
To perform relation generation in a controllable manner, we restrict this procedure in a finite relation space, which is thus reduced to an LLM based relation filtering process (\ie adaptively determining which relations deserve expansion for each entity).  
In terms of entity expansion, we steer LLMs with well-designed progressive prompting augmentation (\ie from coarse-grained to fine-grained prompts), followed by a reliable aggregation with comprehensive consideration of both self-consistency and  semantic relatedness.
To remedy the shortcoming of LLMs in mastering marketing-oriented knowledge, we collect relevant evidence from SupKG and available knowledge bases to empower LLMs in both relation filtering and entity expansion stages.
And then, we specialize in a small and white-box model (dubbed as \emph{Light}{\model}) with a high-quality corpus provided by a strong teacher-LLM involving 175 billion parameters. 
At last, extensive experiments and practical applications in audience targeting verify and justify the effectiveness of the proposed (\emph{Light}){\model}.

\section{Related work}\label{sec:rel}
\subsection{Traditional knowledge graph construction}
Currently, traditional construction process of well-known knowledge bases / KGs usually requires heavy human labor for labeling and evaluation or multiple text mining-based procedures. 
In the line of manual labor-intensive research~\cite{Bollacker2008freebase,Hoffart2013yoga2}, WordNet~\cite{fellbaum98wordnet} is a database that links words with semantic relations, ConceptNet~\cite{conceptnet2017} is a large commonsense knowledge graph consists of a large set of knowledge triples while ATOMIC~\cite{atomic2019} is a crowd-sourced social commonsense KG of inferential knowledge which is organized as typed if-then relations. 
On the other hand, text mining-based approaches usually consist of sub-procedures such as named entity recognition\cite{lample2016ner,ner2022_1,ner2023_1}, relation extraction\cite{Zhu2019re,re2023_1,re2023_2}, semantic parsing\cite{kamath2019sp,sp2022_1,sp2023_1} that require to collect different text corpus to train independent model for each procedure. 
Among them, Quasimodo~\cite{Romero2019quasimodo}, TransOMCS~\cite{Zhang2020transomcs}, DISCOS~\cite{Fang2021DISCOS}, ASCENT~\cite{nguyen2021ascent}, SupKG~\cite{zang2023supkg} construct commonsense or practice-oriented knowledge graphs by making use of linguistic patterns and huge manual processing. 

\paratitle{Open issues.}
These research works usually hinge on collecting different text samples to train multiple independent models for each procedure (\eg named entity recognition and relation extraction) and require huge manual resources for labeling. However, the limited manual resources and text corpora usually restrict the scale of constructed knowledge graph.

\subsection{Language model-driven knowledge graph construction}
For works in this direction, they tend to 1) distill knowledge graphs from LMs, 2) incorporate the power of LMs to enhance knowledge graph construction.
In COMET \cite{comet2019}, implicit knowledge from deep pre-trained language models is transferred to generate explicit knowledge in commonsense knowledge graphs. 
MAMA~\cite{wang2020llmopenkg} proposes to use learned knowledge stored in pre-trained LMs in an unsupervised manner to construct a knowledge graph and compares the quality of it with human-labeled knowledge graph.
BertNet \cite{bertnet2023} generates multiple prompts for KG generation and weights them by the likelihood of prompts given by an LLM.
In LMCRAWL \cite{crawling2023}, relation paraphrases and object paraphrases are utilized for KG generation. 
On the other hands, \cite{milena2023enhancingkg} compares the advanced LLM \eg ChatGPT with the specialized
pre-trained models \eg REBEL for entity and relation
extraction.

\paratitle{Open issues.}
In these research works, they tend to generate knowledge graphs or knowledge bases towards world knowledge / contain knowledge stored in the parameters of LMs, while in industrial scenarios, for example, commercial marketing scenarios, usually pure LMs may be unfamiliar with the marketing-oriented entities and relations hence those approaches cannot meet the requirements.
\section{The Proposed {\model} Framework}
In this section, we dive into the elaboration of the proposed {\model},
a novel \underline{P}rompting \underline{A}ugmented m\underline{I}ning f\underline{R}amework for harvesting marketing-oriented knowledge with LLMs.

\subsection{Formulation}
Essentially, a mining procedure of a knowledge graph could be formulated as follows: with the given source entity $s$, the likelihood of target entity $t$ and its corresponding relation $r$ is calculated as
\begin{equation}
    \begin{split}
        \mathcal{P}(r, t | s) &= \int_{\kappa}\mathcal{P}(\kappa|s)\mathcal{P}(r, t | s, \kappa) \\
        &= \int_{\kappa}\mathcal{P}(\kappa|s) \mathcal{P}(r | s, \kappa) \mathcal{P}(t | s, \kappa, r) \\
    \end{split}
\end{equation}

\begin{itemize}[leftmargin=*]
    \item $\mathcal{P}(\kappa|s)$: \emph{Prior Knowledge}. Intuitively, the knowledge miner commonly struggles to perceive the real meaning of the given entities in the marketing-oriented applications, \eg sneaker-related entities might be preferred with the given entity ``Air''. Hence, the usage of prior knowledge $\kappa$ occupies a critical position in the guidance of knowledge mining.
    \item $\mathcal{P}(r | s, \kappa)$: \emph{Relation Filtering}. Based on the  enormous knowledge $\kappa$, the pure process of relation generation is ideal but generally uncontrollable, especially in marketing-oriented applications where a bundle of relations might be undesired, \eg relation ``Related scientists'' \wrt entity ``The Three Body''.  Therefore, we resort to the powerful capability of LLMs for filtering desired relations $r$ from  the pre-defined relation set $\mathcal{R}$. 
    
    \item $\mathcal{P}(t | s, \kappa, r)$: \emph{Entity Expansion}. With careful prompting engineering, we communicate with LLMs to complete the entity expansion task. However, due to the variety of prior knowledge $\kappa$, coupled with the sensitivity of prompting under slight modifications, a single painstakingly crafted prompt may not always produce the desired results. Instead, we prompt the LLMs for entity expansion with multiple prompting augmentations via a reliable aggregation strategy. 
\end{itemize}
In the following, we will delve into the instantiation of each part with the help of the recent emerging LLMs.

\begin{figure}[!t]
    \centering
    \includegraphics[width=1\columnwidth]{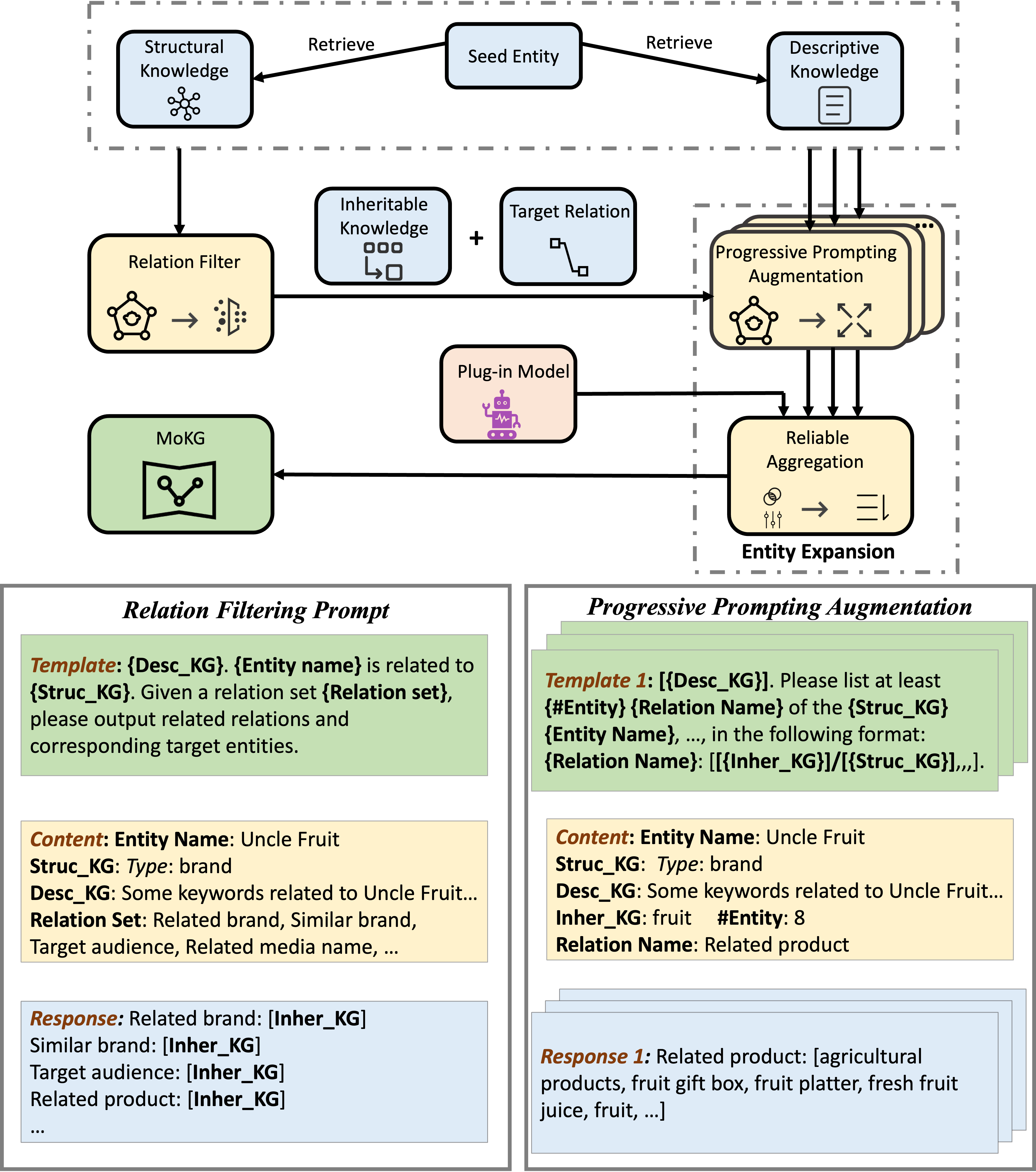}
    \caption{Overall architecture of {\model}.}
    \label{fig:model}
\end{figure}

\subsection{The Usage of Prior Knowledge}
In practice, LLMs commonly struggle to perceive the real meanings of the given entities due to the lack of domain-level knowledge, which severely hinders the capabilities of LLMs in both relation filtering and entity expansion stages. Hence, it is of crucial importance to introduce prior knowledge to specialize in a general LLM for marketing-oriented scenarios. 
In short, for each entity $e$, we mainly collect the following two complementary knowledge to serve as key evidence of LLMs for reasoning.
\begin{itemize}[leftmargin=*]
    \item \emph{Structural Knowledge.} The collection of structural knowledge benefits from the establishment of SupKG~\cite{zang2023supkg}, which involves abundant side information of entities, including brand, category and so on. In order to maintain the reliability of structural knowledge  and introduce as less noise as possible, for each entity, we only collect the one-hop neighbors in the SupKG, coupled with the entity type for disambiguation (\eg The type of entity ``Apple'' could be  ``Brand'' or ``Food'').

    \item \emph{Descriptive Knowledge.} Due to the incomplete and noisy fact of SupKG, we are unable to collect all the convincing evidence for steering LLMs. Hence, we resort to descriptive knowledge from public (\ie Wikipedia) and private knowledge bases. Taking the entity ``The Three Body'' as an example, the SupKG only provides its entity type ``Book'' and its category ``Science Fiction'', while the descriptive summary in Wikipedia offers stronger evidence (\ie author information and brief introduction) to support effective knowledge mining. 
    
\end{itemize}

\subsection{LLM as a Relation Filter}
As mentioned above, to make the whole procedure of relation generation under control, we restrict the process to a finite relation space. Hence a relation filter is encouraged since not all relations in the whole relation set $\mathcal{R}$ are helpful / appropriate for a given entity to expand marketing-oriented entities (\eg given the brand entity ``Apple'', the relation ``Related Food'' may confuse the LLMs for entity expansion). Therefore, we resort to the powerful capability of LLMs for filtering desired relations from the pre-defined relation set $\mathcal{R}$. Given a source entity $s$ with an LLM $\mathcal{M}$, we define the relation filtering as follows:

\begin{equation}
    s \xrightarrow[(1)]{\phi(\cdot;\mathcal{K}), \mathcal{R}} \mathcal{R}^{\mathcal{T}}_s\xrightarrow[(2)]{\mathcal{M}(\cdot ; \rho^{\mathcal{R}}(s, \kappa))}  \mathcal{R}^{\mathcal{F}}_s.
\end{equation}

\begin{enumerate}
    \item \emph{Retrieval}. Considering that the whole relation set $\mathcal{R}$ (includes relations of all entity types in SupKG and marketing experts defined relations for each given entity type) cannot be fed into a single prompt, we first locate the type of the entity by querying the SupKG $\mathcal{K}$, defined as $\phi(\cdot; \mathcal{K})$, \eg $\phi(\text{``Air''};\mathcal{K}) = \text{``Brand''}$. With the entity type\footnote{For simplicity, the retrieved relation set is the same for different entities of the same entity type.},  we would retrieve a sub-relation set $\mathcal{R}^{\mathcal{T}}_s$ from $\mathcal{R}$ for entity $s$, where $|\mathcal{R}^{\mathcal{T}}_s| \ll |\mathcal{R}|$  (\eg ``Related brand'', ``Target audience'' and so on for the Brand ``Air''). 
    \item \emph{Filtering}.  We come up with applying the LLM as a relation filter with structural / descriptive knowledge augmented prompting engineering $\rho^{\mathcal{R}}(s, \kappa)$, \ie denoted as ``\{Struc\_KG\}'' and ``\{Desc\_KG\}'' in Fig.~\ref{fig:model}. With the designed prompt, an LLM would provide 
    i) desired relations $\mathcal{R}_s^{\mathcal{F}}$ that align with the source entity and corresponding prior knowledge, allowing for more reasonable entity expansion;
    ii) a potential target entity corresponding to each desired relation, which could be regarded as a kind of inheritable knowledge for entity expansion. 
\end{enumerate}

\subsection{LLM as an Entity Expansionist}
Due to the variety of prior knowledge, coupled with the sensitivity of prompting under slight modifications, simply employing a single painstakingly crafted prompt may not always produce the desired results. 
To this end, we steer LLMs with careful progressive prompting augmentation, followed by a reliable aggregation that takes both self-consistency and semantic relatedness into consideration.
Given a source entity $s$ and the corresponding relation $r$, a pipeline of the entity expansion with an LLM $\mathcal{M}(\cdot)$ is formulated as follows:
\begin{equation}
    (s,r) \ \ \ \begin{matrix} 
\\ \xrightarrow[]{\mathcal{M}(\cdot ; \rho^{\mathcal{E}}( s, \kappa_1, r))} \mathcal{T}^{1}_{s,r}
\\  \xrightarrow[]{\mathcal{M}(\cdot ; \rho^{\mathcal{E}}( s, \kappa_2, r))} \mathcal{T}^{2}_{s,r}
\\ \xrightarrow[(1)]{\mathcal{M}(\cdot ; \rho^{\mathcal{E}}( s, \kappa_{...}, r))} \mathcal{T}^{...}_{s,r}
\end{matrix}\xrightarrow[(2)]{\mathcal{A}(; )}  \mathcal{T}^{\mathcal{F}}_{s,r}.
\end{equation}

\paratitle{(1) Progressive prompting augmentation.}
It is designed for navigating an LLM $\mathcal{M}$ to different aspects of prior knowledge $\mathcal{\kappa}$, making the expansion procedure more diverse and robust. In particular, we mainly prepare three kinds of knowledge: structural knowledge $\kappa^{\mathcal{S}}$, descriptive knowledge $\kappa^{\mathcal{D}}$ and inheritable knowledge from relation filtering $\kappa^{\mathcal{I}}$, \ie denoted as ``\{Struc\_KG\}'',  ``\{Desc\_KG\}'' and  ``\{Inher\_KG\}'' in Fig.~\ref{fig:model}. By incorporating various knowledge, we progressively construct prompts from coarse-grained to fine-grained manner as $\{\rho^\mathcal{E}(s, r; \kappa^{\mathcal{S}}), \rho^\mathcal{E}(s, r; \kappa^{\mathcal{D}}), 
    \rho^\mathcal{E}(s, r; \kappa^{\mathcal{I}}), \rho^\mathcal{E}(s, r; \kappa^{\mathcal{S}}, \kappa^{\mathcal{D}}), \cdots, \\ \rho^\mathcal{E}(s, r; \kappa^{\mathcal{S}}, \kappa^{\mathcal{D}}, \kappa^{\mathcal{I}})\}$, and  query a  performant LLM repeatedly for stability (Fig.~\ref{fig:model} presents one of the prompts for entity expansion, while more detailed examples of  prompting augmentation will be shown in Sec.~\ref{sec:app}.). As a result, we define the mined entities from  the progressive prompt augmentation as $\left \{ \mathcal{T}^{1}_{s,r}, \mathcal{T}^{2}_{s,r},\cdots, \mathcal{T}^{...}_{s,r} \right \}$. 

\paratitle{(2) Reliable aggregation.}
 After entity expansion with progressive prompting augmentation, we aim at seeking an aggregation mechanism $\mathcal{A}(\cdot)$ in a reliable manner to obtain the final entity expansion result, which takes into consideration the following main aspects:
 \begin{itemize}[leftmargin=*]
     \item \emph{Semantic relatedness} of the knowledge tuple $(s, r, t)$. Here, we plug in a small language model, \ie KG-BERT~\cite{kgbert2019}, which is pre-trained on the hybrid corpus of Wikipedia and SupKG. Subsequently, we tokenize the knowledge tuple $(s, r, t)$ as $\{\text{<}CLS\text{>}, z_1^s, \\ \cdots, z_a^s, \text{<}SEP\text{>}, z_1^r,\cdots,z_b^r, \text{<}SEP\text{>}, z_1^t, \cdots, z_c^t, \text{<}SEP\text{>}\}$, where the 
     source entity is represented as a sentence containing tokens $z_1^s, \cdots, z_a^s$, the relation is represented as a sentence containing tokens $z_1^r, \cdots, z_b^r$ and the target entity is represented as a sentence containing tokens $z_1^t, \cdots, z_c^t$, and feed it into the pre-trained KG-BERT for contextualized representation of each token. We regard the output of token <CLS> as the representation of the knowledge tuple (denoted as $\bm{x}_{s,r,t}$), and the semantic relatedness could be obtained through a linear projection $\text{MLP}(\cdot)$.

     \item \emph{Self-consistency} of the knowledge.  Specifically, it aims at finding the most consistent entities when the LLM performs reasoning in multiple ways. Intuitively, when an entity appears across multiple expansion results, it indicates the higher confidence of LLMs towards the particular target entity, which can then be utilized to reward the highly self-consistent target entities. Here, we denote the total number that $t$ appears across multiple entity expansion procedure as $\sum_{k}\mathbb{I}(t \in \mathcal{T}_{s,r}^k)$, where $k$ indicates a specific round in progressive prompt augmentation and $\mathbb{I}(t \in \mathcal{T}_{s,r}^k)$ returns $1$ if $t$ in $\mathcal{T}_{s,r}^k$ else 0.  
 \end{itemize}

 By integrating both aspects, we define the final score for aggregation as follows:
    \begin{equation}
        \tau_{s,r,t} = \log(1 + \sum_{k}\mathbb{I}(t \in \mathcal{T}_{s,r}^k)) \cdot \text{MLP}(\bm{x}_{s,r,t})
    \end{equation}
    Finally, given a source entity $s$ and relation $r$, we only keep top $K$\footnote{In our experiment, we set $K$ to 8 \wrt the trade-off between multiple metrics.} mined entities with the highest scores from aggregation as the entity expansion result, denoted as $\mathcal{T}^{\mathcal{F}}_{s,r}$.

\begin{figure}[t]
\vspace{-0.4mm}
    \centering
    \includegraphics[width=1.0\columnwidth]{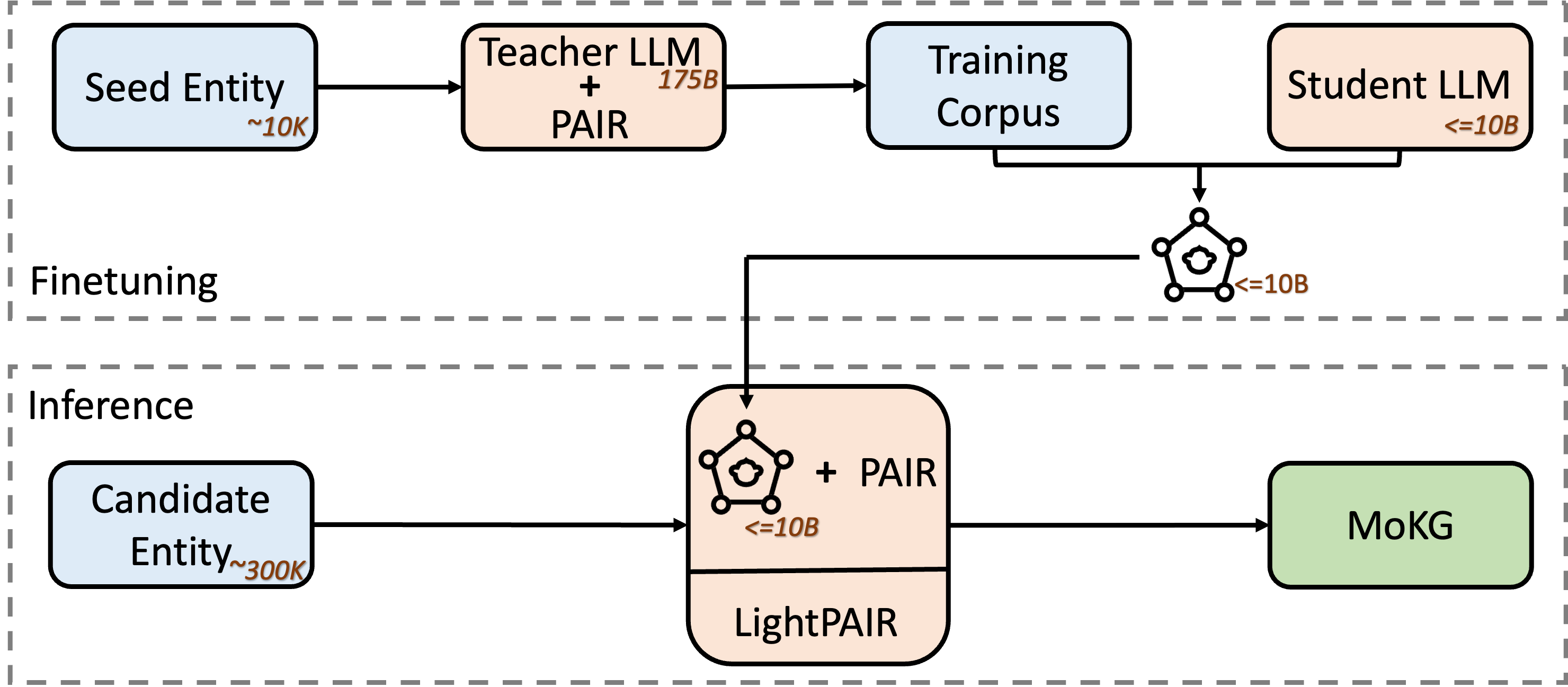}
    \caption{The finetuning and inference pipeline of {\lightmodel}.}
    \label{fig:deployment}
    \vspace{-0.4mm}
\end{figure}

\subsection{{\lightmodel}: Towards Deployment with a Strong Teacher-LLM}
In general, the effectiveness of {\model} greatly hinges on the remarkable advancements of LLMs, whereas, equipping {\model} with today’s state-of-the-art LLMs (\ie GPT 3.5 with 175 billion parameters) is rather difficult even impractical to most of the real-world applications due to far beyond the affordable cost of memory and compute. 
Meanwhile, inspired by the common insight that the corpus generated by a strong teacher-LLM is highly valuable for enhancing smaller task-specific models via careful fine-tuning\cite{hsieh2023distill,shi2023replug,xu2023baize}, we seek to conduct a lightweight and white-box pipeline of {\model} (\ie {\lightmodel}) for large-scale  {\kg} mining from the complete {\model}.
In particular, we collect a group of seed entities (\ie around 4K $\sim$ 10K) from SupKG and feed them into the complete {\model}. Subsequently, we perform a human evaluation (detailed in Sec.\ref{sec:exp_setup}) on the output entity set, followed by bad-case filtering, \eg remove relations / target entities that are annotated as incorrect by human annotators. To obtain a high-quality corpus for fine-tuning the smaller language models, we follow the prompt template of relation filtering and entity expansion to construct questions, coupled with corresponding responses. Based on the {\kg}-oriented corpus, current lightweight LLMs (\eg Bloomz \cite{muennighoff2022bloomz}, GLM \cite{du2022glm}, ChatGLM \cite{zeng2023glmb}, Baichuan-7B~\cite{baichuan2023baichuan2}) could be easily fine-tuned (\ie supervised fine-tuning, short as SFT) via the parameter-efficient Lora~\cite{hu2022lora} strategy.
Finally, we inject the smaller fine-tuned LLM into the relation filtering and entity expansion procedure of {\model}, and adopt the {\lightmodel} for mining large-scale {\kg}.

\subsection{Discussion: Diving into the Progressive Prompting Augmentation}

\paratitle{Merits}
From the specific examples of progressive prompting augmentation presented in Sec.~\ref{sec:app}, we can find that when compared with the non-augmented prompt output, the progressive prompting augmentation has the following virtues: 
\begin{itemize}[leftmargin=*]
\item \textbf{Triggering a more reliable knowledge mining in the industrial setting.} Usually, pure LLMs may be unfamiliar with many marketing-oriented entities and relations such that they tend to hallucinate in their output, \eg "Uncle Fruit power bank" in the case of entity "Uncle Fruit", however, the appropriately injected knowledge can provide the right context for LLMs and steer them to the right reasoning direction, hence the output can be closely related to a brand that sells fruit products.
\item \textbf{Capable of navigating an LLM to different aspects.} Take "Uncle Fruit" as an example, since different knowledge has its emphasis, the output of an LLM \wrt different prior knowledge varies from each other (\eg the output of prompt 1 focuses on common fruit products while the output of prompt 2 focuses more on the product of service) and results in a more diversified output, which we carefully evaluate with two metrics in a coarse-grained to a fine-grained manner and encouraging
results can be found in the “Experiments” part.
\end{itemize}

\paratitle{About time consumption}
Intuitively, the progressive prompting augmentation requires multiple calls to LLMs, incurring more time consumption compared to the original methods. However, this additional time investment is deemed acceptable due to the following three reasons: 
i) The progressive prompting augmentation earns remarkable performance gains compared to original prompting methods, detailed in Table \ref{tab:main-results}.
ii) The knowledge mining task solely relies on offline inference, and there is no requirement for daily inference. As a result, the increased time consumption has no impact on downstream applications.
iii) Since {\lightmodel} is capable of being deployed on a single A100, we can seamlessly accomplish the overall inference task by utilizing multiple GPUs, thereby meeting the time requirements efficiently.

\begin{table*}[t]\centering
\setlength{\tabcolsep}{2.5mm}
{

\caption{Overall performance comparison of different methods towards mining {\kg}. We mark the best and runner-up performance in \textbf{bold} and ``-'' means these methods fail to mine novel entities.} 
\label{tab:main-results}
\begin{tabular}{c|l|c|c|c|c|c|c|c|c}

\toprule
 {} & {}
 & \multicolumn{4}{c|}{{\kg}-181} 
& \multicolumn{4}{c}{{\kg}-500}   \\ 
\cmidrule{3-10}
{} & {Methods} 
 & \multirow{2}{*}{Accuracy} & \multirow{2}{*}{Novelty} & \multicolumn{2}{c|}{Diversity} & \multirow{2}{*}{Accuracy} & \multirow{2}{*}{Novelty} & \multicolumn{2}{c}{Diversity}   \\ 
 {} & {} & {} & {} & {AEE} & {ILAD}  & {} & {} & {AEE} & {ILAD} \\
\midrule

  \multirow{2}{*}{KG Completion} & {BERT} 
& {58.4\%} & {-} & {43.0} & {4.86} 
& {57.7\%} & {-} & {42.7}  & {4.97}  \\
\cmidrule{2-10} 

 {} & {TRMP} 
& {\textbf{91.1\%}} & {-} & {13.8} & {3.12} 
& {\textbf{91.3\%}} & {-} & {14.1} & {3.07}  \\
\midrule
 \multirow{2}{*}{KG  Construction} & {LMCRAWL} 
& {86.3\%} & {\textbf{41.2\%}} & {36.3} & {\textbf{5.98}} 
& {85.2\%} & {\textbf{41.7\%}} & {37.1} & {\textbf{6.00}} \\

\cmidrule{2-10}
 {} & {COMET} 
& {86.7\%} & {35.9\%} & {26.1}  &5.06 
& {85.9\%} & {34.6\%} & {25.3}  & 4.95 \\

\midrule

 \multirow{4}{*}{OURS} & \cellcolor[gray]{0.8}{\model} 
&\cellcolor[gray]{0.8} {\textbf{90.1\%}} & \cellcolor[gray]{0.8}{\textbf{40.4\%}} & \cellcolor[gray]{0.8}{\textbf{43.7}}  & \cellcolor[gray]{0.8}{5.88} 
& \cellcolor[gray]{0.8}{\textbf{90.7\%}} & \cellcolor[gray]{0.8}{\textbf{43.6\%}} & \cellcolor[gray]{0.8}{\textbf{42.8}}   & \cellcolor[gray]{0.8}{5.78}  \\

\cmidrule{2-10}
 {} & {\quad - Agg} 
& {88.7\%} & {39.6\%} & {30.8}  & {5.89} 
& {88.9\%} & {36.4\%} & {31.3}   & {5.84}  \\

\cmidrule{2-10}
 {} & {\quad - Agg \& Pr} 
& {86.9\%} & {36.8\%} & {30.8}  & {5.82} 
& {87.2\%} & {34.2\%} & {31.4}  & {5.87}  \\

\cmidrule{2-10}
 {} & {\quad - Agg \& Pr \& Rf} 
& {84.9\%} & {39.2\%} & {\textbf{46.3}}  &{\textbf{6.15}} 
& {84.3\%} & {39.4\%} & {\textbf{47.2}}  & {\textbf{6.20}}  \\

\bottomrule
\end{tabular}
}
\end{table*}

\section{Experiments}\label{sec:exp}
\subsection{Experimental Setup}\label{sec:exp_setup}
\subsubsection{Datasets}

Since {\model} aims to mine diverse marketing-oriented knowledge using a group of seed entities and a candidate relation set, we randomly collect two seed entity sets from the well-established SupKG. These entity sets consist of 181 entities (\ie  {\kg}-181) and 500 entities (\ie  {\kg}-500). With domain knowledge, we meticulously gather 105 relations (\eg ``Related Food'', ``Issue Voucher'', ``Prize of Food'' and ``Brand of Product'') as the candidate relation set for entity expansion.

\subsubsection{Baselines and Variants}
We mainly compare {\model} with the following  baselines and variants: 
\begin{itemize}[leftmargin=*]
    \item \emph{KG Completion based}, which attempts to mine knowledge in the established SupKG through textual matching and structural matching:
i) \textbf{BERT~\cite{devlin2019bert}}, which completes an existing knowledge graph by textual matching using semantic-based representation.
ii) \textbf{TRMP~\cite{yang2023interested}}, which constructs a knowledge graph by retrieval and ranking of entity pairs in a progressive manner and ensembles the entity representation across time to enhance the knowledge graph stability.

    \item \emph{KG Construction based}, which aims at exploring commonsense or open-world knowledge with deep pre-trained language models:  i) \textbf{COMET~\cite{comet2019}}, which learns from existing knowledge bases and generates novel nodes and edges in commonsense descriptive natural languages. ii) \textbf{LMCRAWL~\cite{crawling2023}}, which designs multiple prompts to mine a knowledge graph out of a large language model (GPT 3.5 in this paper) by following the procedure of Subject Paraphrasing $\rightarrow$ Relation Generation $\rightarrow$ Relation Paraphrasing $\rightarrow$ DK Object Generation.
    
    \item \emph{Variants of} {\model}: To further examine the merits of each well-designed part in {\model}, we prepare three variants:
    i) \textbf{{\model} -Agg,} which discards the \underline{Agg}regation operation and it is equivalent to the most performant version of Progressive prompting.
    ii) \textbf{{\model} -Agg \& Pr,} which discards the \underline{Agg}regation operation and the \underline{Pr}ogressive prompting and it is equivalent to directly utilize the output of Relation Filtering.
    iii) \textbf{{\model} -Agg \& Pr \& Rf,} which discards the \underline{Agg}regation operation, the \underline{Pr}ogressive prompting and the \underline{R}elation \underline{f}iltering and it is equivalent to directly utilize an LLM for the task of knowledge mining.

\end{itemize}

For {\model}, we equip it with today’s state-of-the-art LLM (\ie GPT 3.5 with 175 billion parameters). 
During the progressive prompting augmentation process, we call the LLM for each specific prompt for 3 times. For reliable aggregation, the KG-BERT~\cite{kgbert2019} consists of 110 million parameters, which is the same as BERT~\cite{devlin2019bert}.

\subsubsection{Evaluation Protocols and Metrics}
We mainly evaluate the quality of the knowledge mining with human annotation, where three annotators are involved using a correctness / incorrectness judge for each extracted knowledge tuple. A tuple is considered ``accepted'' if at least two annotators deem it to be true knowledge, and ``rejected'' if at least two annotators vote it as false. To ensure the credibility of human evaluation, we initially gather all predicted tuples from both baselines and our method. These tuples are then randomly shuffled before being presented for human annotation. In this way, the annotators are unaware of which model a specific knowledge tuple is derived from, which helps to ensure the fairness in the evaluation process.

Overall, we comprehensively measure the quality of mined knowledge from the following aspects:
i) \emph{Accuracy} refers to the proportion of accepted tuples.
ii) \emph{Novelty} refers to the proportion of extracted entities that do not appear in the SupKG.
iii) \emph{Diversity} is evaluated with two metrics from coarse-grained to fine-grained manner. 
\emph{AEE} (Average Entity Expansion) refers to the average number of target entities expanded from seed entities. 
\emph{ILAD} (Intra-List Average Distance)~\cite{wu2022survey} is a distance-based diversity metric,  calculating the average pair-wise (Euclidean) distance among all the target entities given a source entity in representation space\footnote{The representations are derived from BERT that are pre-trained on Wikipedia and SupKG.}. 
In general, when evaluating the quality of mined knowledge, the accuracy metric comes first. With  accuracy guarantee, we hope the mined knowledge with higher novelty and diversity thus better facilitating downstream tasks in real-world environment.

\subsection{Performance Overview}\label{sec:exp_overview}

In this part, we report the evaluation results of different knowledge mining approaches for mining {\kg} in Table \ref{tab:main-results} and the major findings can be summarized as follows:
\begin{itemize}[leftmargin=*]
\item \textbf{(Superior performance)} The proposed {\model} consistently achieves remarkable performance across all datasets and key metrics, indicating its effectiveness in mining marketing-oriented knowledge graphs. Besides, beyond its considerable performance in accuracy metric, its superior performance in novelty and different aspects of diversity further indicates that the harvested knowledge graph can provide considerable complementary knowledge to the original knowledge graph, \ie SupKG, thus further facilitating broader applications to downstream tasks in real-world environments.
\item \textbf{(Compared with KG completion based methods)} It is worthwhile to note that KG completion based methods, especially TRMP, yield impressive performance \wrt the accuracy metric. However, these methods fail to offer extra knowledge since they fail to mine novel entities, which is unacceptable for the mining of {\kg} in real-world applications.
\item \textbf{(Compared with KG construction based methods)} Not surprisingly, KG construction based methods (\ie LMCRAWL and COMET) generally bring a considerable amount of novel knowledge with the help of deep pre-trained language models. However, the proposed {\model} still surpasses them, attributed to the following well-designed components tailored for mining {\kg}: adaptive relation filtering and progressive prompting augmentation, empowered by domain-level knowledge (More case studies are presented in Section~\ref{exp:casestudy1}).
\item \textbf{(Ablation studies of {\model})} Clearly, the performance of {\model} would drop a lot once the corresponding module is discarded, indicating that all of our elaborate designs occupy critical positions in mining high-quality and high-diversity {\kg} for online marketing. It is noted that the variant {\model} -Agg \& Pr \& Rf (namely basic prompting method which means directly utilizing an LLM) ranks highest \wrt the diversity metric, demonstrating the ability of a performant LLM in finding potential target entities for knowledge graph construction. However, directly utilizing an LLM for this task results in a rapid decrease of accuracy and novelty as compared to {\model} and its other variants, which further justifies our model design.
\end{itemize}

\begin{table*}[!t]\centering
\setlength{\tabcolsep}{2mm}
{

\caption{Performance evaluation of {\lightmodel} instantiated by different student LLMs. We mark the best and runner-up performance of student LLMs in \textbf{bold}.} 
\label{tab:sft-results}
\begin{tabular}{c|c|c|c|c|c|c|c|c|c}

\toprule
 \multicolumn{3}{c|}{Models}  & \cellcolor[gray]{0.8}{\model} & \multicolumn{6}{c}{\lightmodel}  \\ 
   \cmidrule{1-10}
\multicolumn{3}{c|}{LLMs(Size) }
& \cellcolor[gray]{0.8}{GPT 3.5(175B)} 
& {GLM(10B)} 
& {GLM(10B)}
& {Bloomz(7B)}
& {ChatGLM2(6B)}
& {Baichuan2(7B)}
& {Qwen2(7B)}
\\
 \cmidrule{1-10}
\multicolumn{3}{c|}{SFT ?}
& \cellcolor[gray]{0.8}{-} 
& {\XSolidBrush} 
& {\CheckmarkBold}
& {\CheckmarkBold}
& {\CheckmarkBold}
& {\CheckmarkBold}
& {\CheckmarkBold}
\\ 
\cmidrule{1-10}
\midrule
{} & \multicolumn{2}{c|}{Accuracy} & \cellcolor[gray]{0.8}{90.1\%} & {79.2\% }& {\textbf{89.0\%} }& {80.8\% }& {86.3\%}& {\textbf{90.3\% }}& {80.6\% } \\ 
\cmidrule{2-10}
{Train-} & \multicolumn{2}{c|}{Novelty} & \cellcolor[gray]{0.8}{40.4\%} & {\textbf{39.9\%}}& {31.0\%}& {29.0\%}& {28.8\%}& {\textbf{31.5\%}}& {25.8\%} \\ 
\cmidrule{2-10}
{25K} & {\multirow{2}{*}{Diversity}} & {AEE} & \cellcolor[gray]{0.8}{43.7} & {27.44}& {35.71}& {\textbf{48.47} }& {39.21}& {\textbf{41.08}}& {25.02} \\   
\cmidrule{3-10}

{} & {} & {ILAD} & \cellcolor[gray]{0.8}{5.88} & {\textbf{6.14}}& {5.77}& {\textbf{6.12}}& {5.82}& {5.96}& {5.74} \\ 
\cmidrule{1-10}
\midrule
{} & \multicolumn{2}{c|}{Accuracy} & \cellcolor[gray]{0.8}{90.1\%} & {79.2\% }& {\textbf{90.7\%} }& {82.7\%}& {88.9\%}& {\textbf{92.2\%}}& {80.6\%} \\ 
\cmidrule{2-10}
{Train-} & \multicolumn{2}{c|}{Novelty} & \cellcolor[gray]{0.8}{40.4\%} & {\textbf{39.9\%}}& {31.5\%}& {30.2\%}& {29.5\%}& {\textbf{34.3\%}}& {29.1\%} \\ 
\cmidrule{2-10}
{100K} & {\multirow{2}{*}{Diversity}} & {AEE} & \cellcolor[gray]{0.8}{43.7} & {27.44}& {37.10}& {\textbf{48.03} }& {38.94}& {\textbf{41.13}}& {25.89} \\ 
\cmidrule{3-10}
{} & {} & {ILAD} & \cellcolor[gray]{0.8}{5.88} & {
\textbf{6.14}}& {5.83}& {\textbf{6.14} }& {5.88}& {5.79}& {5.49} \\ 
\bottomrule
\end{tabular}
}
\end{table*}

\subsection{Further Probe}

\subsubsection{{\lightmodel} with Different Student-LLMs}
In this part, we investigate into the effectiveness of {\lightmodel}, augmented by different student-LLMs.

\paratitle{Data collection for SFT.}
With the help of GPT 3.5, we generate two datasets for fine-tuning the smaller LLMs towards relation filtering and entity expansion, consisting of 25K (\ie Train-25K) and 100K (Train-100K) samples 
constructed from (4K-10K) seed entities
, respectively. In detail, we carefully choose various types of entities and relations from SupKG and utilize the teacher LLM (GPT 3.5) to do relation filtering, and entity expansion by progressive prompting and reliable aggregation. Subsequently, we select 3 annotators to perform a thorough manual verification of the results derived from the teacher LLM, which ensures the creation of a high-quality training dataset. 
On the other hands, we utilize the  {\kg}-181 as the evaluation dataset in this experiments
~\footnote{Due to the space limit, we omit the experimental results on {\kg}-500 in this paper. However, it is worth mentioning that similar trends are observed as well.}.

\paratitle{Experimental settings of student-LLMs.}
With the collected training corpus derived from the strong teacher-LLM, we adopt several currently popular open-source LLMs to instantiate the {\lightmodel}, \ie GLM~\cite{du2022glm}, Bloomz~\cite{muennighoff2022bloomz} and ChatGLM2~\cite{zeng2023glmb}, Baichuan-7B~\cite{baichuan2023baichuan2}, Qwen-7B~\cite{qwen2023qwen}, and set the optimizer, maximum context size, batch size, and learning rate to Adam, 4096, 8 and 5e-5 respectively.

\paratitle{Results and analysis.}
From  Table \ref{tab:sft-results}, by comprehensively taking multiple metrics into consideration we could find that {\lightmodel} with GLM(10B) and Bainchuan2(7B) could achieve comparable performance of the complete {\model} with GPT 3.5(175B), implying that high-quality corpus generated by a strong teacher-LLM can greatly enhance smaller task-specific models. Besides, as the training corpus accumulates, an increasing trend among multiple metrics can be observed which also depicts the significance of collecting high-quality task-oriented corpus for enhancing task-specific models. When compared with those student-LLMs with SFT (short for Supervised Fine-Tuning), though a pure student-LLM without SFT ranks highest \wrt ILAD, the lowest accuracy of the collected knowledge prohibits its application in real-world industrial environments. 

\begin{figure}[t]
\vspace{-0.75em}
    \centering
    \includegraphics[width=1\columnwidth]{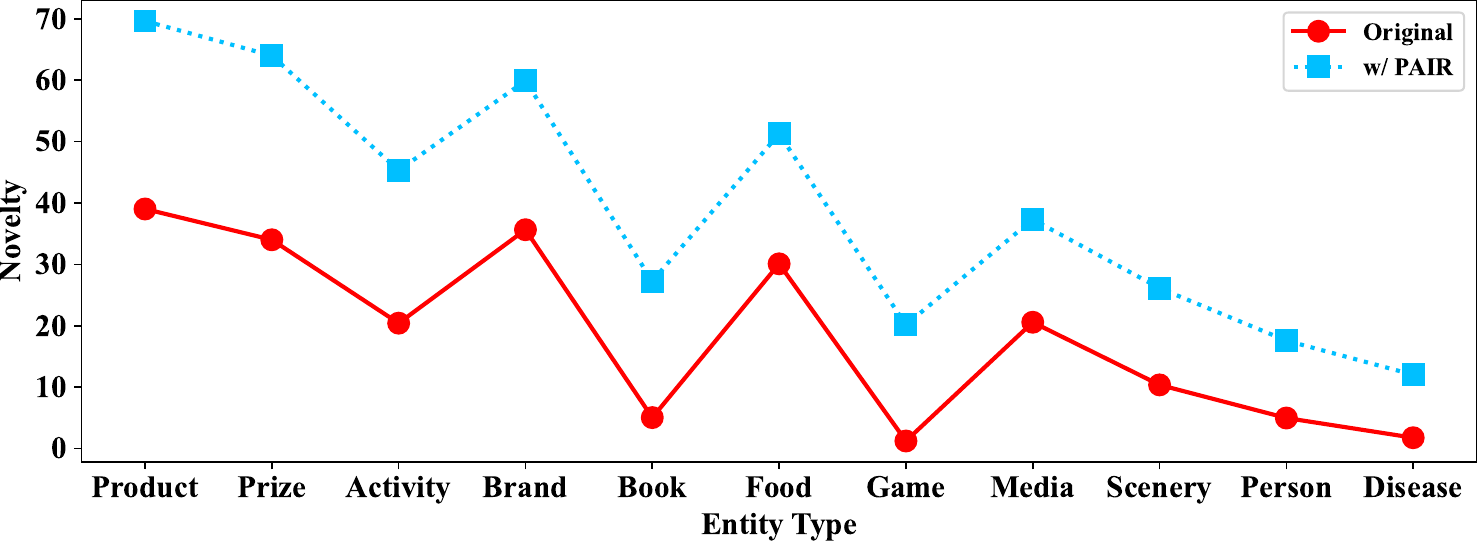}
    \caption{Illustration of the knowledge graph enriched by the {\model}. }
    \label{fig:diversity_comparison}
    \vspace{-0.75em}
\end{figure}

\begin{table*}[t]\centering
\setlength{\tabcolsep}{1.75mm}
{

\caption{Case study towards the usage of prior knowledge in {\model}. We \textbf{bold} the hallucinative entities and fallacious entities in \textbf{\textcolor{red}{red}} and \textbf{\textcolor{blue}{blue}}, respectively.} 
\label{tab:case_study}
\begin{tabular}{l|l|l|l}

\toprule
 {Source entity}& {Relation} &{type}  & {Target entities}    \\ 
\cmidrule{1-4}

{Mi Xiao Quan } & \multirow{2}{*}{related media name} & {w/o knowledge} & \textbf{\textcolor{blue}{Journey to the West (\begin{CJK}{UTF8}{gbsn}西游记\end{CJK})}}\\ \cmidrule{3-4} 

(\begin{CJK}{UTF8}{gbsn}米小圈上学记\end{CJK}) &&  \cellcolor[gray]{0.8}{w/ knowledge} & \cellcolor[gray]{0.8}{Tom and Jerry(\begin{CJK}{UTF8}{gbsn}猫和老鼠\end{CJK}), Boonie Bears(\begin{CJK}{UTF8}{gbsn}熊出没\end{CJK})} \\ 
\midrule 
{CKA} & \multirow{2}{*}{target audience} & {w/o knowledge} & \textbf{\textcolor{blue}{System Administrator (\begin{CJK}{UTF8}{gbsn}系统管理员\end{CJK})}}\\ \cmidrule{3-4}
(\begin{CJK}{UTF8}{gbsn}空手道协会\end{CJK}) & &  \cellcolor[gray]{0.8}{w/ knowledge} & \cellcolor[gray]{0.8}{Karate Enthusiasts(\begin{CJK}{UTF8}{gbsn}空手道迷\end{CJK}), Wushu Master(\begin{CJK}{UTF8}{gbsn}武术师\end{CJK})} \\ 
\midrule 
{Uncle Fruit } & \multirow{2}{*}{related brand} & w/o knowledge & \textbf{\textcolor{red}{Fruit Education(\begin{CJK}{UTF8}{gbsn}果儿教育\end{CJK})}}, \textbf{\textcolor{blue}{Canon(\begin{CJK}{UTF8}{gbsn}佳能\end{CJK})}}\\ \cmidrule{3-4} 
(\begin{CJK}{UTF8}{gbsn}果叔\end{CJK}) & &  \cellcolor[gray]{0.8}{w/ knowledge}&  \cellcolor[gray]{0.8}{Xianfeng Fruit(\begin{CJK}{UTF8}{gbsn}鲜丰水果\end{CJK}), Fruitday(\begin{CJK}{UTF8}{gbsn}天天果园\end{CJK})} \\ 
\midrule 
{The Three Body } & \multirow{2}{*}{similar movie} & w/o knowledge & The Wandering Earth(\begin{CJK}{UTF8}{gbsn}流浪地球\end{CJK}) \\ \cmidrule{3-4}
(\begin{CJK}{UTF8}{gbsn}三体\end{CJK}) & &   \cellcolor[gray]{0.8}{w/ knowledge} & \cellcolor[gray]{0.8}{Interstellar(\begin{CJK}{UTF8}{gbsn}星际穿越\end{CJK}), Star Trek(\begin{CJK}{UTF8}{gbsn}星际迷航\end{CJK})} \\ 
\midrule 
{Gas Coupon} & \multirow{2}{*}{product of prize} & w/o knowledge & Fuel Card (\begin{CJK}{UTF8}{gbsn}加油卡\end{CJK}) \\ \cmidrule{3-4}
(\begin{CJK}{UTF8}{gbsn}加油券\end{CJK}) & &  \cellcolor[gray]{0.8}{w/ knowledge} & \cellcolor[gray]{0.8}{Diesel(\begin{CJK}{UTF8}{gbsn}柴油\end{CJK}), Gasoline(\begin{CJK}{UTF8}{gbsn}汽油\end{CJK}), Gas Gift Card(\begin{CJK}{UTF8}{gbsn}加油礼品卡\end{CJK})}\\ 
\midrule 

{Tuxi Living Plus} & \multirow{2}{*}{related company} & w/o knowledge& \textbf{\textcolor{red}{Tuxi Catering(\begin{CJK}{UTF8}{gbsn}兔喜餐饮\end{CJK})}}\\ \cmidrule{3-4}
(\begin{CJK}{UTF8}{gbsn}兔喜生活+\end{CJK}) & &  \cellcolor[gray]{0.8}{w/ knowledge}&  \cellcolor[gray]{0.8}{Carrefour(\begin{CJK}{UTF8}{gbsn}家乐福\end{CJK}), CR Vanguard(\begin{CJK}{UTF8}{gbsn}华润\end{CJK}), Walmart(\begin{CJK}{UTF8}{gbsn}沃尔玛\end{CJK})}\\ 
\bottomrule
\end{tabular}
}
\end{table*}

\subsubsection{Case Study I: How the Prior Knowledge Improves {\model}}\label{exp:casestudy1}
In this section, we concentrate on the effectiveness of prior knowledge towards {\model}, which is expected to serve as strong evidence for relation filtering and entity expansion. We present the results of entity expansion for PAIR w/ knowledge and PAIR w/o knowledge in Table~\ref{tab:case_study}, respectively. 
Clearly, we find that target entities produced by the complete {\model} are more related to the corresponding source entities. On the contrary, {\model} w/o knowledge usually mine  hallucinative entities (\eg ``Fruit Education'' and ``Tuxi Catering'') and fallacious
entities (\eg ``Canon'' with the entity ``Uncle Fruit'' and relation ``related brand''), severely hindering the quality of {\kg}.

\subsubsection{Case Study II: How the Original Knowledge Graph is Enriched with {\model}}\label{exp:casestudy2}
Since several efforts have been made for connecting users and various services / products in Alipay through hierarchical and spatio-related relations, {\model} could provide another complementary knowledge, \ie marketing-related knowledge. Here, we take a close look at the novelty at the entity level by randomly selecting several specific entity types and presenting the average novelty in the original knowledge and \kg with {\model} in Fig.~\ref{fig:diversity_comparison}.
We find that the original knowledge graph is greatly enriched by {\model}, especially for the entity types  ``Book'', ``Game'' and ``Disease'', which are rather rare before.

\begin{figure}[t]
    \centering
    \includegraphics[width=1\columnwidth]{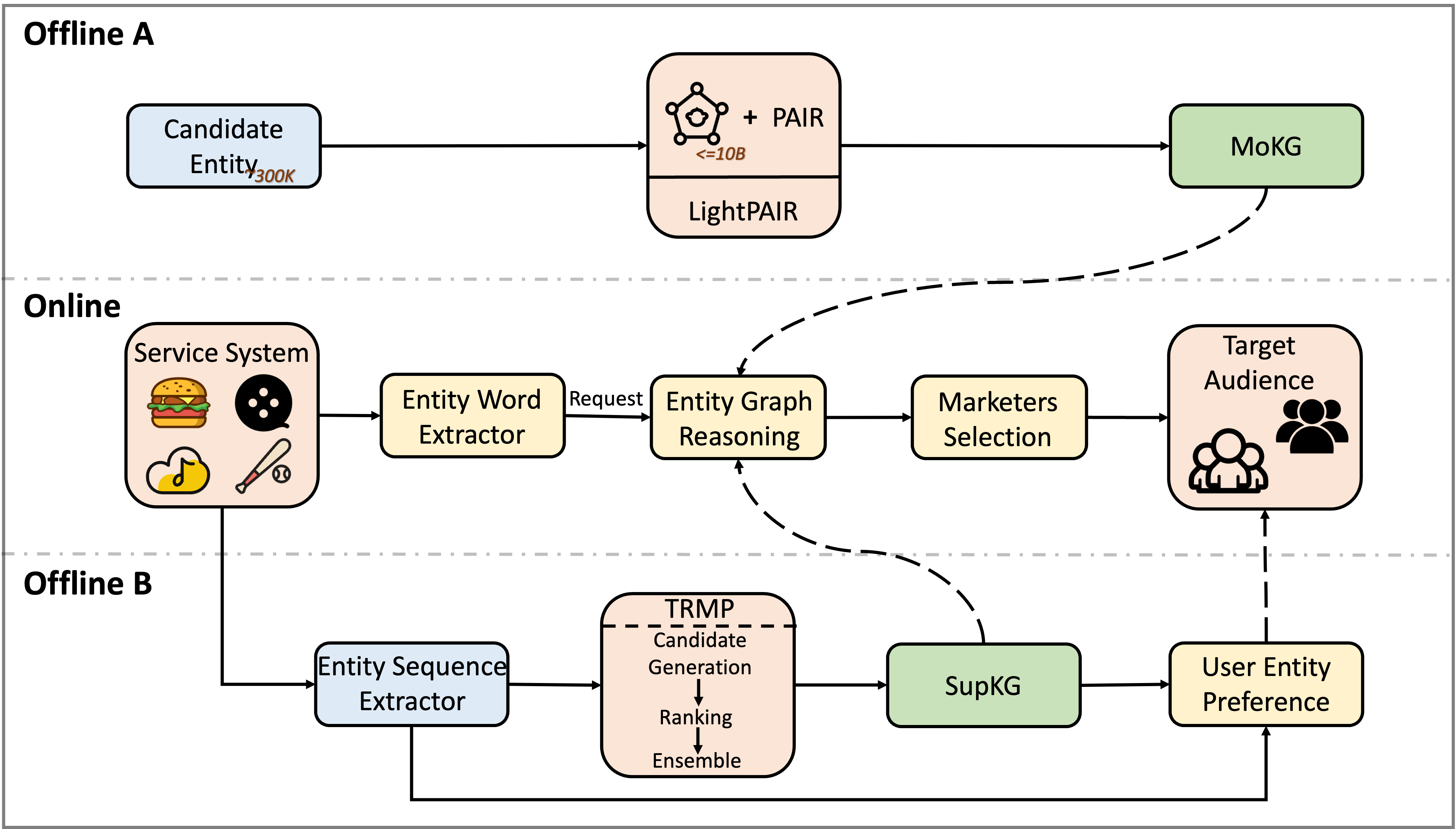}
    \caption{Deployment of {\lightmodel} (\ie ``Offline A'') for audience targeting and its comparison to the original EGL system with  TRMP framework~\cite{yang2023interested}(\ie ``Offline B'').}
    \label{fig:application}
\end{figure}

\begin{table}[t]
\vspace{-0.5mm}
  \setlength{\tabcolsep}{4mm}
{
\caption{Performance comparison for audience targeting with the metric \# TAC (thousand). ``RI'' means the relative improvement of {\lightmodel} \wrt EGL.}
  \centering
  
    \begin{tabular}{cccc}
      \hline
      Scenarios &{ EGL }&{{\lightmodel}} &{RI} \\
      \hline
      { Uncle Fruit}  & 7*1  & 8*7 & +15.3\%  \\
      { The Three Body} & 3*3 & 3**6 & +98.1\%  \\ 
       { Schwarzkopf } & 2*7 & 4*9  & +93.1\% \\ 
      { Biscuits voucher} & 1*2 & 1*3  & +31.2\% \\ 
      { Land Lords}  & 9*2 & 2**2  & +122.0\% \\ 
      { Gas Coupon} & 3*8 & 6*1  & +89.2\% \\ 
   
    \hline
    
\end{tabular}
\label{tab:metrics}
}
\vspace{-0.5mm}
\end{table}

\subsubsection{Application: Audience Targeting}
At last, we will demystify the practical application of {\kg} in audience targeting, one of the fundamental issues of conducting marketing campaigns~\cite{zhuang2020hubble}, which aims at effectively and efficiently locating target audiences who are interested in certain products / services / campaigns. As shown in Fig.~\ref{fig:application}, similar to the original EGL system with TRMP framework~\cite{yang2023interested}, the deployment of {\lightmodel} is decoupled with the online serving, thus satisfying the requirements of scalability and timeliness in industrial environments. 
We report the performance of {\lightmodel} and EGL based audience targeting in Table \ref{tab:metrics} with the metrics \# TAC (\ie the number of \underline{T}arget \underline{A}udiences \underline{C}overed). In Table~\ref{tab:metrics}, each scenario could be abstracted as several entities. 
Compared to the deployed EGL system, {\lightmodel} could achieve performance gains over the EGL by 15\% - 122\% in different marketing scenarios \wrt the number of Target Audiences Covered.
Both of the observations show the effectiveness and efficiency of the proposed {\lightmodel} towards practical online marketing.

\section{Conclusion}
In this paper, we propose a novel {\model} and its lightweight version {\lightmodel} for harvesting marketing-oriented knowledge with LLMs, which are equipped with LLM based adaptive relation filtering and progressive  prompting augmentation based entity expansion 
followed by a reliable aggregation that takes both self-consistency and semantic relatedness into consideration. Towards lightweight online serving, we specialize in a small and white-box (\emph{Light}){\model}, which is fine-tuned with high-quality corpora provided by a strong teacher-LLM. Extensive experiments 
demonstrate the effectiveness of the proposed (\emph{Light}){\model} via several
key metrics for knowledge graph evaluation and practical applications in audience targeting also verify its applicability in real-world marketing scenarios. In the future, we would like to equip the proposed (\emph{Light}){\model} with the metapath-oriented entity expansion such that the expansion process can be more controllable and explainable.
\clearpage
\appendix
\section{Examples of Progressive Prompting Augmentation}\label{sec:app}
To further demonstrate the usage of different aspects of prior knowledge in the entity expansion procedure, we present a series of specific examples of progressive prompting augmentation in Figure 6 - 8, where different types of knowledge is distinguished by color, \ie \textcolor{blue}{Struc\_KG},  \textcolor{red}{Desc\_KG} and  \textcolor{green}{Inher\_KG}.

\begin{figure}[h]
    \centering
    \includegraphics[width=1.07\columnwidth]{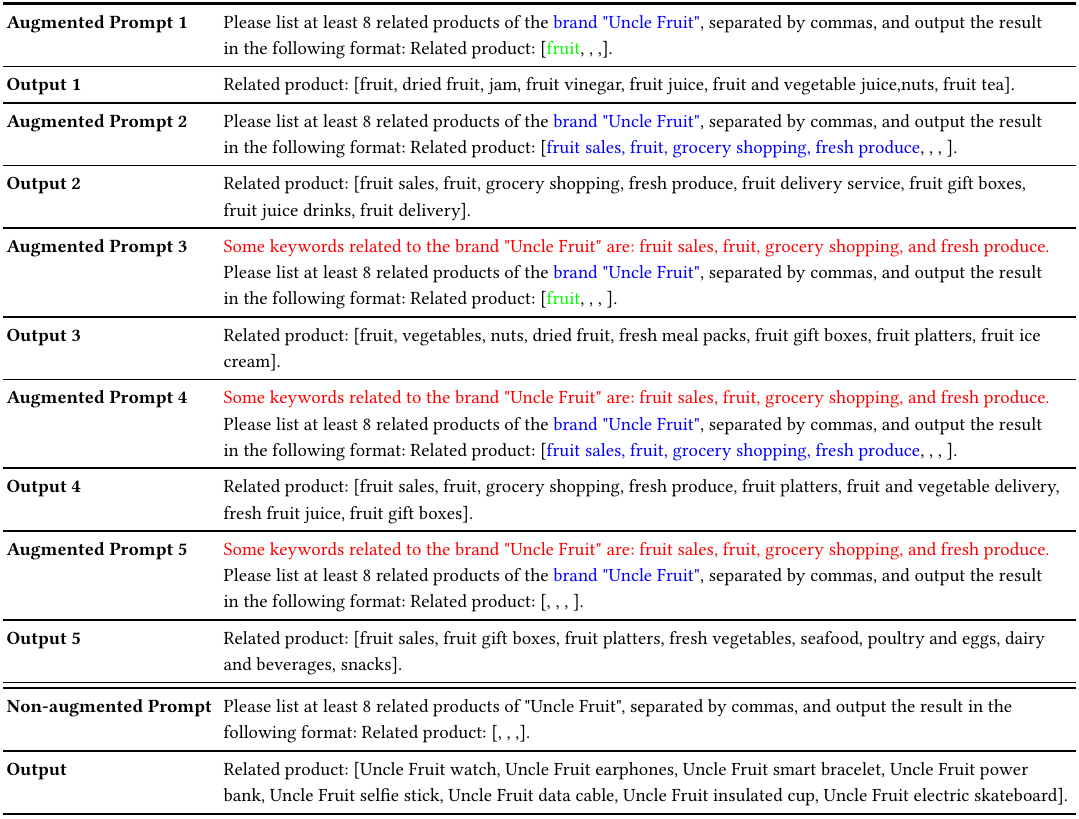}
    \caption{An example (source entity: Uncle Fruit, relation: related product) of calling an LLM to carry out progressive prompting augmentation.}
    \label{fig:eg1}
\end{figure}

\begin{figure}[h]
    \centering
    \includegraphics[width=1.07\columnwidth]{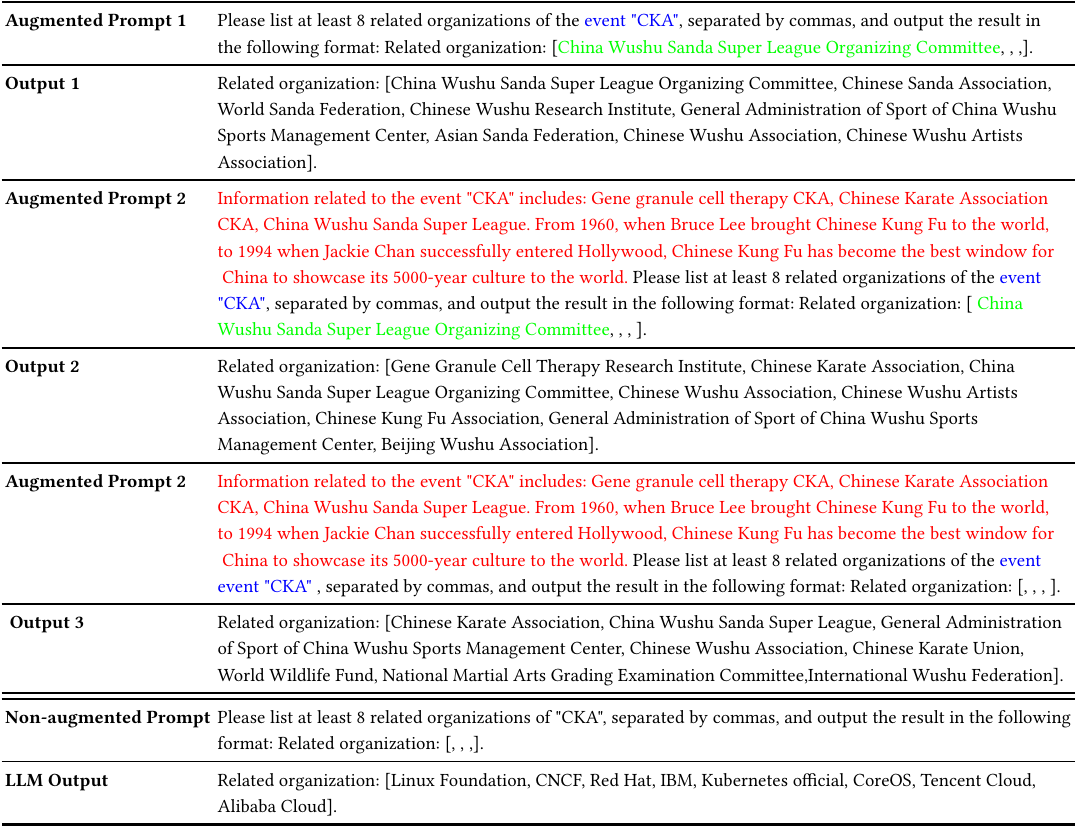}
    \caption{An example (source entity: CKA, relation: related organization) of calling an LLM to carry out progressive prompting augmentation.}
    \label{fig:eg2}
\end{figure}

\begin{figure}[h]
    \centering
    \includegraphics[width=1\columnwidth]{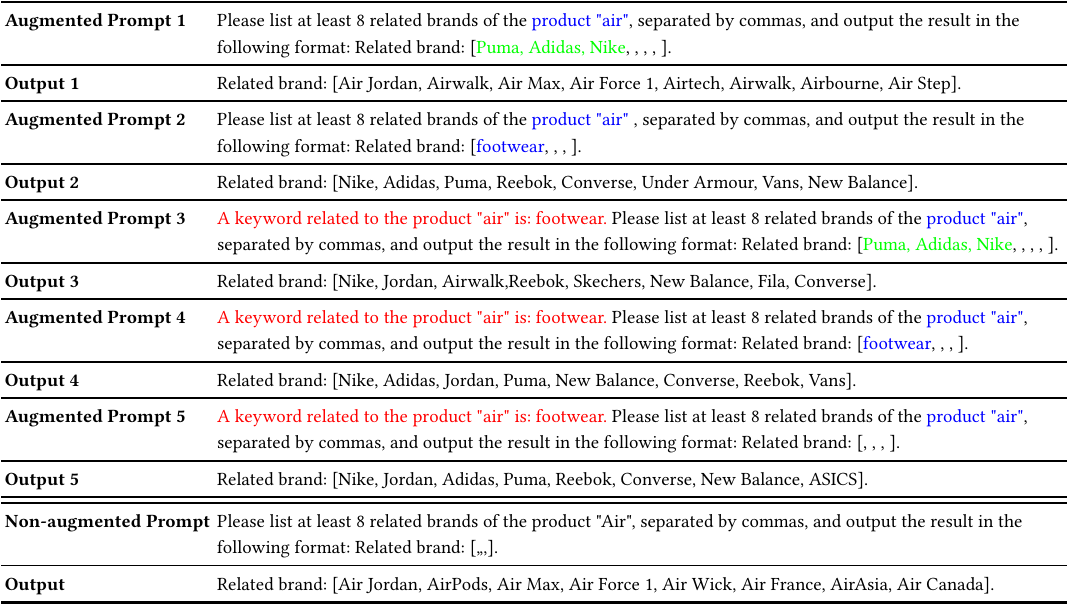}
    \caption{An example (source entity: air, relation: related brand) of calling an LLM to carry out progressive prompting augmentation.}
    \label{fig:eg3}
\end{figure}

\bibliographystyle{ACM-Reference-Format}
\bibliography{references}

\end{document}